\documentclass{article}
\pdfoutput=1
\PassOptionsToPackage{numbers, compress}{natbib}

\usepackage{epigraph}

\usepackage[nonatbib, preprint]{nips_2018}

\usepackage[utf8]{inputenc} 
\usepackage[T1]{fontenc}    
\usepackage{hyperref}       
\usepackage{url}            
\usepackage{booktabs}       
\usepackage{amsfonts}       
\usepackage{nicefrac}       
\usepackage{microtype}      

\usepackage{amsmath}
\usepackage{graphicx}
\usepackage{algorithm}
\usepackage[noend]{algorithmic}
\usepackage{fancybox}
\usepackage{wrapfig}
\usepackage{amsfonts, amsmath, amssymb, bbm}
\usepackage{graphics}                 
\usepackage[usenames,dvipsnames]{color}
\DeclareGraphicsExtensions{.pdf,.gif,.jpg} 

\usepackage[font=small,labelfont=bf,tableposition=top]{caption}

\DeclareCaptionLabelFormat{andtable}{#1~#2  \&  \tablename~\thetable}

\usepackage{natbib}

\newcommand{\vv}[1]{\mathbf{#1}}
\newcommand{\bb}[1]{\mathbb{#1}}

\newcommand{\R}{\mathbb{R}}
\newcommand{\N}{\mathcal{N}}

\title{Disentangled VAE Representations for Multi-Aspect and Missing Data}

\author{
    Samuel K.~Ainsworth\thanks{\url{http://samlikes.pizza}} \\
    Computer Science and Engineering\\
    University of Washington\\
    \texttt{skainsworth@gmail.com}\\
    \And
    Nicholas J.~Foti\\
    Computer Science and Engineering\\
    University of Washington\\
    \texttt{nfoti@uw.edu}\\
    \And
    Emily B.~Fox\\
    Computer Science and Engineering\\
    University of Washington\\
    \texttt{ebfox@uw.edu}\\
}

\begin{document}

\maketitle

\begin{abstract}
Many problems in machine learning and related application areas are fundamentally variants of conditional modeling and sampling across multi-aspect data, either multi-view, multi-modal, or simply multi-group. For example, sampling from the distribution of English sentences conditioned on a given French sentence or sampling audio waveforms conditioned on a given piece of text.
Central to many of these problems is the issue of missing data: we can observe many English, French, or German sentences individually but only occasionally do we have data for a sentence pair. Motivated by these applications and inspired by recent progress in variational autoencoders for grouped data~\citep{ainsworth2018interpretable}, we develop factVAE, a deep generative model capable of handling multi-aspect data, robust to missing observations, and with a prior that encourages disentanglement between the groups and the latent dimensions.
The effectiveness of factVAE is demonstrated on a variety of rich real-world datasets, including motion capture poses and pictures of faces captured from varying poses and perspectives.
\end{abstract}


\section{Introduction}

Developing generative models for complex data is an important task in modern machine learning. Deep generative models address this problem by using deep neural networks to parameterize the distribution of the data. These methods have seen great success in a variety of settings~\cite{shrivastava2017learning, kingma2013auto}.  In this paper, we focus on \emph{multi-aspect data}, consisting of multiple views, modalities, or other definitions of grouped observations.  For example, imagine we have text in multiple languages; datasets with text, images, and categorical responses; or data collected from multiple, potentially heterogeneous sensors.  Building deep generative models for such multi-aspect data still presents significant challenges.

We tackle this problem within the context of variational autoencoders~\cite{kingma2013auto}. These methods define complex encoder and decoder networks that map high-dimensional observations to a low-dimensional space, and then back again.  To learn these complex maps requires a tremendous amount of data.  Additionally, each observation must be fully observed at both training and test time.  Often, our large (or even small) datasets have significant missingness.  For our multi-aspect data, we assume missingness in the form of individual datapoints missing certain aspects (view, modality, etc.).  For example, a sensor drops out, a translation is unavailable, or a modality is not present.  The VAE is simply not applicable in such situations as it treats the collection of aspects as one large input.  Instead, previous work has focused on imputing the missing values apriori~\cite{vedantam2017generative}; however, two-stage approaches are statistically suboptimal, and especially problematic in the structured missing setting we focus on. 


We aim to leverage explicit correlations between the aspects to extend VAEs to handle multi-aspect data, and also allow us to utilize any available data.  In particular, we propose a \emph{factorization} of the encoder and decoder mappings over the aspects, and refer to the resulting model as \emph{factVAE}.  Our specification encourages each aspect to manifest itself on a sparse subset of the latent dimensions.  The two critical insights are (i) a means of coherently combining aspect-specific encoders and (ii) ensuring consistent support on the latent space of each aspect-specific encoder/decoder pair. For the former, we leverage a product of experts framework for combining encoders inspired by the method of~\cite{vedantam2017generative}.  The result is a variational distribution on the latent space whose entropy decreases as more aspects are included.  For the latter, we incorporate shared sparsity on the weights defining the output of the encoder and input of the decoder; the sparsity is encouraged through a group-sparse prior~\cite{xu2015bayesian}.

Side benefits of our sparsity include being able to better handle limited amounts of data and inferring interpretable relationships between latent space activations and aspects, as in~\cite{ainsworth2018interpretable}. Additionally, the resulting sparsity pattern can be used to inform us of which aspects are most informative in capturing dimensions of variability in our data. Fundamentally, the proposed framework yields a \emph{disentangled} latent space where each aspect is linked to a distinct set of latent components.  Correlations between aspects are captured by the overlapping supports of the latent representations.
  
We demonstrate factVAE on a variety of datasets. First, we analyze motion capture sequences, where aspects correspond to groups of joints forming limbs and the core. Next, we explore an image dataset consisting of multiple views of people's heads, with each view corresponding to an aspect.  Both datasets have a limited number of observations, a challenge for traditional VAE methods.  Additionally, we simulate significant missing data by removing aspects (limbs for mocap and views for the image dataset), making traditional VAEs ill-suited.  (The image dataset has naturally occurring missingness, as well.)  Our method provides state-of-the-art reconstruction performance in these settings while also yielding interpretable latent spaces. The resulting model provides a unified and robust means for handling many types of multi-aspect data, even in the presence of potential missingness.

\section{factVAE: Disentangling VAEs with factorized mappings}

We start with the standard VAE formulation of an observation $\vv{x} \in \R^{D}$ embedded into a $K$-dimensional latent representation $\vv{z} \in \R^{K}$~\cite{kingma2013auto}.  The VAE consists of two components, an encoder referred to as the \emph{inference network} and a decoder referred to as the \emph{generator}.  The inference network provides a mapping from a given observation to a variational distribution 
\begin{align}
q(\vv{z}\mid \vv{x}) = \mathcal{N}(\mu_{\phi}(\vv{x}),\Sigma_{\phi}(\vv{x}))
\end{align}
on the latent space.  Here, both the mean $\mu_\phi(\cdot)$ and (diagonal) covariance $\Sigma_{\phi}(\vv{x})$ are defined using deep neural networks. The generator provides a mapping from a latent code, $\vv{z}$, to a distribution on observations defined as:
\begin{align}
\label{eq:vae_factor_model}
	\vv{z} &\sim \N(\vv{0}, \vv{I}) \\
	\vv{x} &\sim \N(f_\theta (\vv{z}), \vv{D}(\vv{z})).
\end{align}
Here, the mean and (diagonal) covariance of the generator's distribution on observations, $f_\theta(\cdot)$ and $\vv{D}_\theta(\cdot)$, respectively, are specified via deep neural networks with parameters $\theta$.  

When we are faced with multi-aspect data---which could be a collection of multimodal data sources, multiple views, or data naturally decomposing into groups of observations---the VAE treats all dimensions jointly, attempting to a learn a complex inference network and generator oblivious to the underlying structure.  Recently, the output interpretable VAE (oi-VAE)~\cite{ainsworth2018interpretable} was proposed to handle such grouped data and leverage within-group correlation structure and between-group sparse dependencies.  One of the key goals is also to uncover interpretable relationships between the dimensions of the latent code and the observation groups.  In particular, each latent dimension generates a sparse subset of the observation groups. 

Formally, the oi-VAE is specified as follows. Write $\vv{x}$ as $[\vv{x}^{(1)},\ldots,\vv{x}^{(G)}]$ for some $G$ groups.  The oi-VAE defines \emph{group-specific generators} as follows:
\begin{align}
\vv{z} &\sim \mathcal{N}(\vv{0}, \vv{I}) \\
\vv{x}^{(g)} &\sim \N(f^{(g)}_{\theta_g} (\mathbf{W}^{(g)} \textbf{z}),\vv{D}^{(g)}).
\end{align}
Here, $\vv{W}^{(g)} \in \R^{p\times K}$ is a group-specific linear transformation between the latent representation $\vv{z}$ and the group generator $f_{\theta_g}^{(g)}$.  Critically, the latent representation $\vv{z}$ is shared over all the group-specific generators.  A benefit of this formulation---beyond supporting group-specific generators---is that one can interpret the relationships between group-specific activations through the latent representation, just as in a standard linear latent factor model.  To further aid in interpretability, and to better handle limited data scenarios (a situation that typically plagues standard VAEs), the oi-VAE specification places a sparsity-inducing prior~\cite{Kyung:2010} on the columns of the latent-to-group matrix $\mathbf{W}^{(g)}$.  When the $j$th column of the weight matrix, $\vv{W}_{\cdot,j}^{(g)}$, is all zeros then the $j$th latent dimension, $\vv{z}_j$, will have no influence on group $g$.  
In order to avoid learning small latent-to-group weights $\vv{W}^{(g)}$ only to be re-amplified by downstream network layers, a standard normal prior is also placed on the parameters of each generative network, $\theta_g \sim \mathcal{N}(\vv{0}, \vv{I})$.

Although the oi-VAE focuses the generator on group-structured observations, providing both interpretability and an ability to handle more limited data scenarios, the framework cannot directly handle multimodal data sources or missing groups of observations.  In particular, the inference network is the \emph{same} as in the standard VAE, treating all observations jointly.  For multimodal data, one could imagine leveraging architectures deployed in other neural network situations, such as combining modality-specific features extracted with an appropriate neural network model~\cite{ngiam2011multimodal,srivastava2012multimodal}.  However, this entangles all of the groups into all of the latent dimensions, making it hard to distinguish which dimensions encode which modalities. Furthermore, this approach still cannot handle missing groups of observations.  (Note that the oi-VAE \emph{generator} could straightforwardly handle multimodal data by defining different likelihoods on different groups.)
\begin{figure}[t!]
\vspace{-0.1in}
    \centering
\begin{tabular}{ccc}
	\includegraphics[height=0.9in]{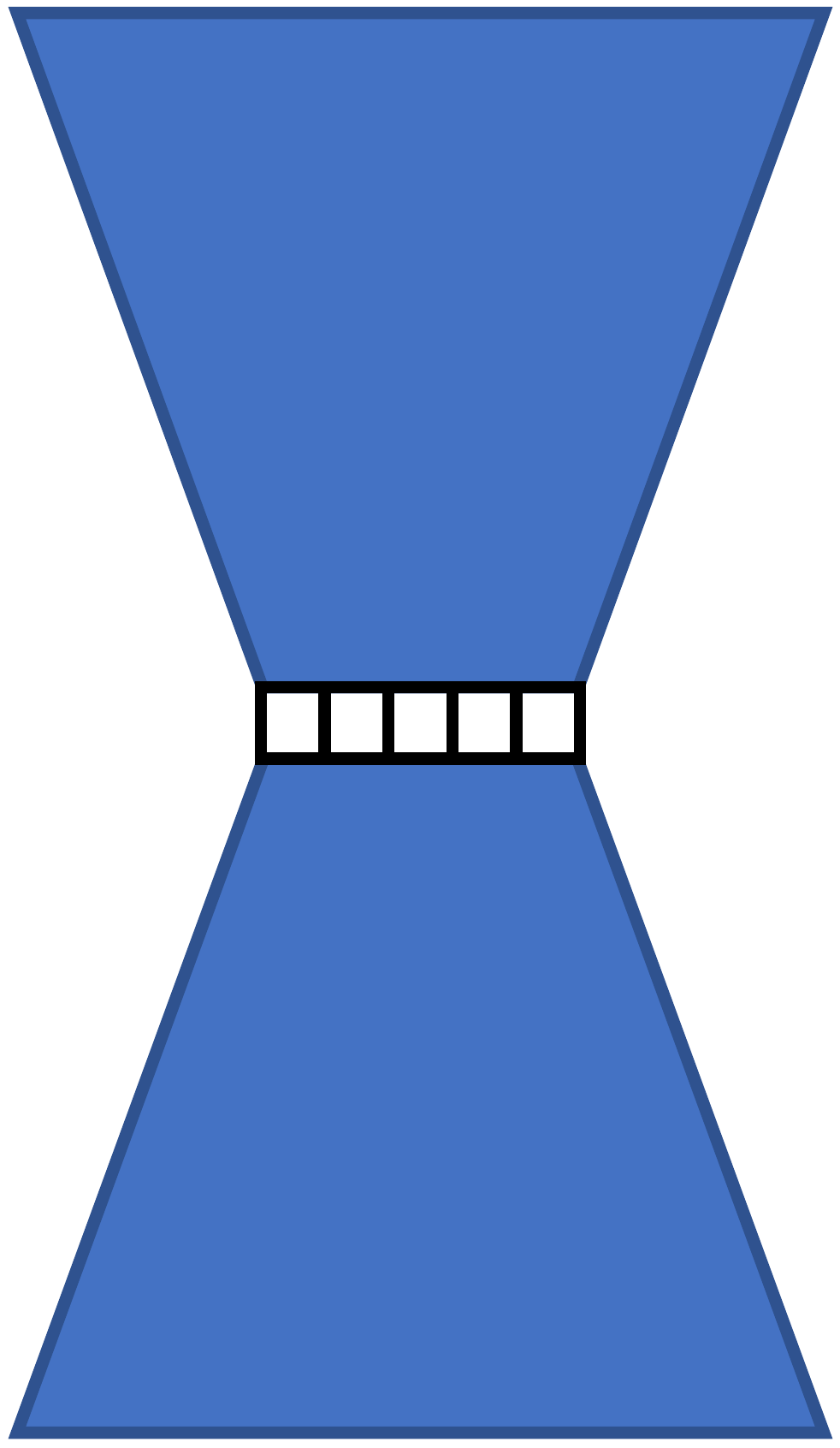} & \includegraphics[height=0.9in]{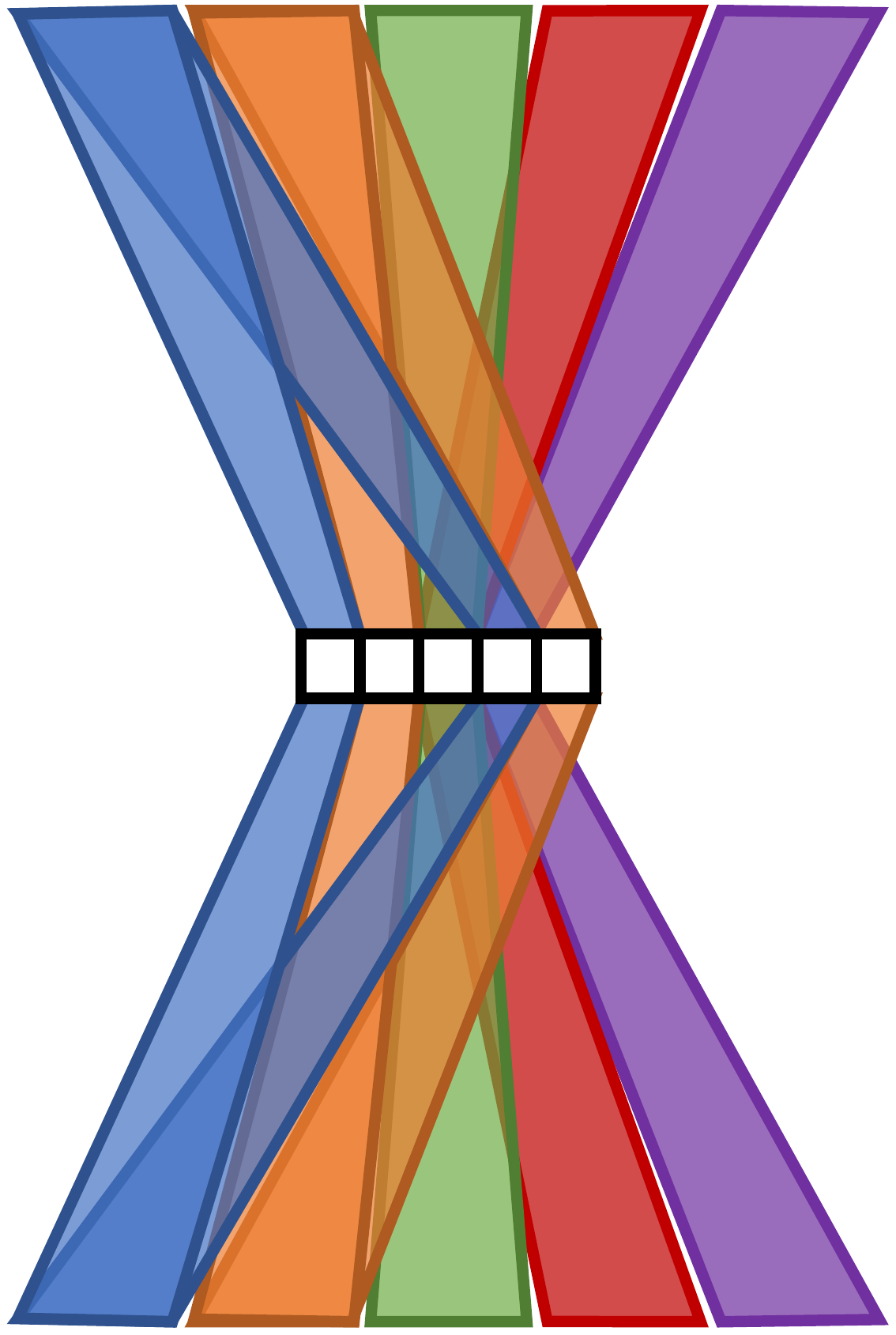} &
	\includegraphics[height=0.9in, width=3.88in]{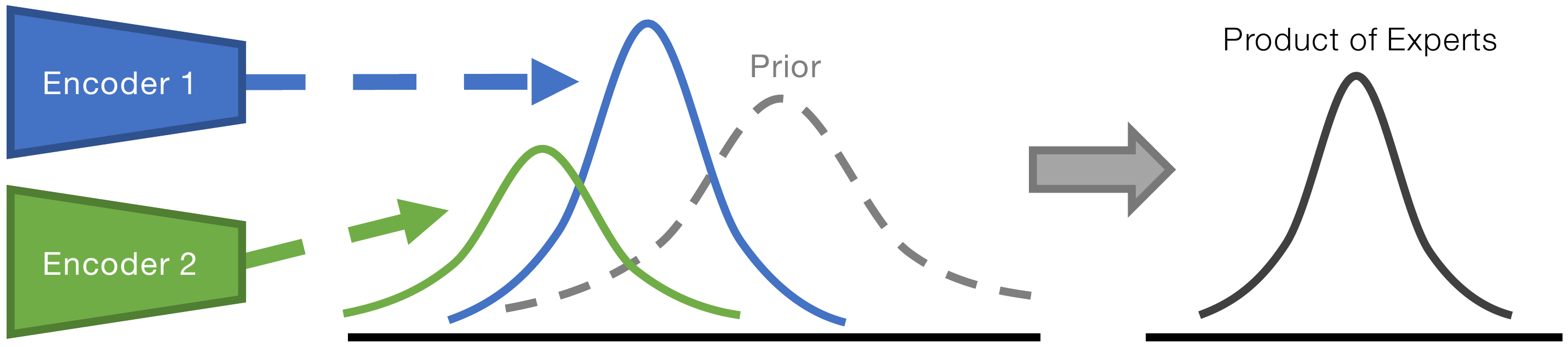}
\end{tabular}
\caption{\small Inference network and generator bottleneck cartoon for the standard VAE (\emph{left}) and factVAE (\emph{middle}), where the factVAE decomposes both the inference and generative networks across observation groups with each sparsely mapping to and from a set of latent dimensions. \emph{Right:} Combining group-specific encoders with a product of experts to form the overall variational distribution.}
\vspace{-10pt}
\label{fig:highlevel}
\end{figure}

We propose a fully \emph{factorized VAE} (factVAE) that considers a group-wise factorization of the inference network, as well.  FactVAE fully supports inference with missing aspects by utilizing an inference network that flexibly aggregates approximate posteriors only over the available groups. The most efficient information flow occurs when a given observation group only influences the dimensions of the variational distribution corresponding to those responsible for generating that observed group.  (We assume the latter is sparse, following the oi-VAE generator specification.)  See Fig.~\ref{fig:highlevel}.  Two critical questions remain:
\begin{enumerate}
	\item \emph{How do we combine these group-specific inference networks into a coherent variational distribution?}
	\item \emph{How do we encourage consistent sparsity patterns to appear on the inference and generator sides?}
\end{enumerate}
These two questions must be answered jointly, since any design choice addressing 1. must be very careful not to violate the sparse group-latent component relationship constraints from 2. It is worth noting that averaging on parameters (or equivalently layer activations) does \textit{not} satisfy these criteria. Even if there are sparse connections between the groups and the parameters, the sparsity will introduce zeros into the average producing some influence on latent components that should be independent of a group, violating 2. For factVAE, we apply a product of experts (PoE) formulation in which each group assumes its own inference network, outputting its own approximate posterior $q_g(\vv{z} | \vv{x}^{(g)})$. These individual variational distributions are aggregated by taking their product,
\begin{equation}
q(\vv{z} | \vv{x}) = p(\vv{z}) \prod_{g \in \mathcal{O}} q_g(\vv{z} | \vv{x}^{(g)}).
\end{equation}
See Fig.~\ref{fig:highlevel}(\emph{right}).  In particular we choose normal distributions for $p(\vv{z})$ and $q_g(\vv{z} | \vv{x}^{(g)})$, allowing us to compute the $q(\vv{z} | \vv{x})$ in closed form.\footnote{The closed form solution is given by $\Lambda = \Sigma^{-1}= I + \sum_{g\in\mathcal{O}} \Lambda^{(g)}$ and $\mu = \Lambda^{-1} \sum_{g\in\mathcal{O}} \Lambda^{(g)} \mu^{(g)}$.} A product of normal experts has the convenient property that its entropy only decreases as more observations are included, meaning that our uncertainty monotonically decreases the more we observe. This is in contrast to other approaches which promote the opposite behavior. Our choice here is related to that made in~\cite{vedantam2017generative}. We further compare and contrast our formulation with alternatives in Section \ref{section:related-work}.

Sparsity is respected by parameterizing each $q_g(\vv{z} | \vv{x}^{(g)})$ by a mean $\mu^{(g)}_\theta(\vv{x}^{(g)})$ and diagonal precision matrix $\Lambda^{(g)}_\theta(\vv{x}^{(g)})$. By enforcing sparsity in $\Lambda^{(g)}_\theta(\vv{x}^{(g)})$ latent components that are independent of the group $g$ have infinite variance in $q_g(\vv{z} | \vv{x}^{(g)})$.  That is, this group is completely uninformative of that latent dimension. We define $\Lambda^{(g)}_\theta(\vv{x}^{(g)})=\text{diag}(|\vv{V}^{(g)} \varphi_\theta^{(g)}(\vv{x}^{(g)})|)$ where $\vv{V}^{(g)}\in\bb{R}^{K \times H}$ and $\varphi_\theta^{(g)}(\vv{x}^{(g)})\in\bb{R}^H$. To have consistent encoder/decoder sparsity, we enforce the column sparsity pattern from $\vv{W}^{(g)}$ to be the same as the row sparsity on $\vv{V}^{(g)}$. To this end, we stack $\vv{V}^{(g)}$ and $\vv{W}^{(g)}$ to produce $\vv{\Phi}^{(g)}=[\vv{W}^{(g)\top}, \vv{V}^{(g)}]^\top$ and then apply group sparsity prior on the columns of $\vv{\Phi}^{(g)}$. This allows us to simultaneously learn the sparsity pattern across both the inference and generative networks, allowing the model to learn which groups and latent components should interact throughout training.

Many Bayesian sparsity-inducing priors exist throughout the literature~\citep{xu2017bayesian, carvalho2009handling}. A popular class of them are global-local shrinkage priors. Despite the flexibility of many of these priors, they are not amenable to fast variational inference and do not recover exact zeros. Instead we use a hierarchical Bayesian \textit{group-lasso} prior on the columns of $\vv{\Phi}^{(g)}$ in order to encourage entire columns to be shrunk to zero. The prior takes the following form~\cite{Kyung:2010}:
\begin{align}
\gamma_{gj}^2 &\sim \text{Gamma}\left(\frac{p + 1}{2}, \frac{\lambda^2}{2}\right) \\
\vv{\Phi}^{(g)}_{\boldsymbol{\cdot}, j} &\sim \mathcal{N}(\vv{0}, \gamma_{gj}^2 \vv{I})
\end{align}
where Gamma($\cdot,\cdot$) is defined by shape and rate, and $p$ denotes the number of rows in each $\vv{\Phi}^{(g)}$. The rate parameter $\lambda$ defines the amount of sparsity, with larger $\lambda$ implying more sparsity in the relationships between latent components and groups. Marginalizing over $\gamma_{gj}^2$ induces group sparsity over the columns of $\vv{\Phi}^{(g)}$; the MAP of the resulting posterior is equivalent to a group lasso penalized objective.

\subsection{Optimizing the factVAE}
\begin{algorithm}[tb]
   \caption{Collapsed SVI for factVAE}
   \label{alg:vi}
\begin{algorithmic}
   \STATE {\bfseries Input:} data $\mathbf{x}_1, \dots, \mathbf{x}_N$, sparsity parameter $\lambda$
   \STATE Let $\tilde{\mathcal{L}}$ be $\mathcal{L}(\theta, \Phi)$ but without $- \lambda \sum_{g,j} || \mathbf{\Phi}^{(g)}_{\boldsymbol{\cdot}, j} ||_2$.
   \REPEAT
   \STATE Pick some $\mathbf{x}_i$ with its subset of available groups $\mathcal{O}$.
   \STATE Randomly select some subset of groups $\mathcal{I} \subset \mathcal{O}$. Construct $q(\vv{z} | \vv{x}_i) = p(\vv{z}) \prod_{g\in \mathcal{I}} q_g(\vv{z} | \vv{x}_i^{(g)})$.
   \STATE Calculate $\nabla_\theta \tilde{\mathcal{L}}$, and $\nabla_\Phi \tilde{\mathcal{L}}$, evaluating the likelihood on all available groups $\mathcal{G}$.
   \STATE Update $\theta$ with an optimizer of your choice.
   \STATE Let $\Phi_{t+1} = \Phi_{t} - \eta \nabla_\Phi \tilde{\mathcal{L}}$.
   \FORALL{groups $g$, and $j=1$ {\bfseries to} $K$}
       \STATE Set $\vv{\Phi}^{(g)}_{\boldsymbol{\cdot}, j} \gets \frac{\mathbf{\Phi}^{(g)}_{\boldsymbol{\cdot}, j}}{||\mathbf{\Phi}^{(g)}_{\boldsymbol{\cdot}, j}||_2} \left(||\mathbf{\Phi}^{(g)}_{\boldsymbol{\cdot}, j}||_2 - \eta \lambda \right)_+ $
   \ENDFOR
   \UNTIL{convergence in both $\hat{\mathcal{L}}$ and $- \lambda \sum_{g,j} || \mathbf{\Phi}^{(g)}_{\boldsymbol{\cdot}, j} ||_2$}
\end{algorithmic}
\end{algorithm}
We train the factVAE by adapting standard stochastic variational inference for VAEs to handle missing data. At training time we use the available groups to produce individual approximate posteriors which are aggregated through the product of experts formulation. With an estimate $q(\vv{z} | \vv{x})$ in hand, we can sample through the model, producing a likelihood $p(\vv{x} | \vv{z})$. Finally we can marginalize out groups that are unobserved, and evaluate the likelihood only on the observed groups. 

More specifically, we deploy collapsed variational inference over the scale parameters $\gamma$,
\begin{equation}
\begin{split}
\log\ p(\mathbf{x}) &= \log \int p(\mathbf{x} | \mathbf{z}, \Phi, \theta) p(\mathbf{z}) p(\Phi | \gamma^2) p(\gamma^2) p(\theta) \,d\gamma^2 \,dz \\
&= \log \int \left( \int p(\Phi, \gamma^2) \,d\gamma^2 \right) \frac{p(\mathbf{x} | \mathbf{z}, \Phi, \theta) p(\mathbf{z}) p(\theta)}{q_\theta(\mathbf{z} | \mathbf{x}) / q_\theta(\mathbf{z} | \mathbf{x})} \,dz \\
&\geq \mathbb{E}_{q_\theta(\mathbf{z} | \mathbf{x})} \left[ \log p(x | \mathbf{z}, \Phi, \theta) \right]  - \mathbb{KL}(q_\theta(\mathbf{z} | \mathbf{x}) || p(\mathbf{z})) + \log p(\theta) - \lambda \sum_{g,j} || \vv{\Phi}^{(g)}_{\boldsymbol{\cdot}, j} ||_2 \\
&\triangleq \mathcal{L}(\theta, \Phi),
\end{split}
\end{equation}
with the subtle difference that we have absorbed the inference networks' parameters (previously $\phi$) into $\theta$ since the sparsity prior is shared across both parameters in inference and generative networks.

To optimize $\mathcal{L}(\theta, \Phi)$ we alternately perform stochastic gradient steps on $\theta$ and proximal gradient descent on $\Phi$, since our hierarchical Bayesian prior admits an efficient proximal operator~\citep{ainsworth2018interpretable,Parikh:2013}. See Algorithm \ref{alg:vi} for pseudocode.

In our training of factVAE, for all of our experiments, we also handicap the inference network by randomly dropping out groups to promote robustness in the presence of missing data even when the dataset itself is complete. This has the added benefit of encouraging the model to learn to sample \textit{across} groups, as opposed to simply performing joint reconstruction.\footnote{We will release code upon acceptance.}

\section{Related work}
\label{section:related-work}
Although there are some previous works investigating deep generative models for multi-aspect data, to our knowledge none learn a disentangled representation through sparsity or offer any results with more than a couple of groups (typically two). The conditional VAE (CVAE)~\citep{sohn2015learning} considered extending the VAE framework to handle conditional distributions between fixed input and output domains. As such, the CVAE only supports conditional sampling from the input to the output domains. However, recent work has begun to tackle the issue of modeling the joint distribution.

The joint multimodal VAE (JMVAE)~\citep{suzuki2016joint} effectively endows each group with its own inference network and combines them through a mixture distribution with uniform weights. 
The central idea is to learn two separate VAEs that share a latent representation and then use their individual inference networks at test time based on whichever group is present. The authors mention that extending the model to more than two groups is possible, although we found that it does require some adjustment. In particular, it is not clear how to reconstruct when a subset of groups are available, as opposed to just one. The Appendix of~\citep{vedantam2017generative} notes that the JMVAE model is equivalent to a uniform mixture distribution over the individual approximate posteriors: $q(\vv{z} | \vv{x}) = \frac{1}{|\mathcal{O}|}\sum_{g\in\mathcal{O}} q_g(\vv{z} | \vv{x}^{(g)})$. We refer to this slight reinterpretation of the model as ``JMVAE+''. We also investigated learning the mixture weights, but did not find that it produced any meaningful benefit. The JMVAE+ interpretation indicates an approach to handling inference with multiple available groups, but it brings along its own issues. 

First, a mixture of Gaussians can only increase in entropy as more components are added, meaning that the JMVAE becomes \textit{less} certain as it receives more data. This is in contrast with factVAE which only ever decreases its uncertainty with the addition of new information. Furthermore, the mixture distribution indicates that a reconstruction from the model must be based on information from only one input modality. Therefore, JMVAE is incapable of fusing information across groups. Finally, there is no closed form for the Kullback–Leibler divergence that appears in the corresponding ELBO, requiring an additional lower bound. 

A clever approach was taken in~\citep{vedantam2017generative} for learning visual models of both images $\vv{x}$ and their binary features $\vv{y}$, eg. \texttt{wearing\_hat}. \citep{vedantam2017generative} also applied a product of experts (PoE) posterior approximation but only for the binary features. In total it had three distinct inference networks: $q(\vv{z} | \vv{x})$, $q(\vv{z} | \vv{y})$, and $q_g(\vv{z} | \vv{x}, \vv{y})$. While tractable for two groups, this means that extending the work to handle more than two groups would require an exponential blowup in the number of inference networks. For our Faceback example in Section \ref{sec:faceback}, that would necessitate 127 networks! We take inspiration from their application of PoE for aggregating approximate posteriors from the inference networks, but we apply it across all modalities obviating the need for a combinatorial number of inference networks. 
In addition, we support missingness and emphasize sparsity in the learned representation.

An orthogonal direction was considered with the output intepretable VAE (oi-VAE)~\citep{ainsworth2018interpretable}. oi-VAE aimed to model grouped observations with sparse, interpretable group-latent component interactions but without considering missing or multimodal data. Specifically oi-VAE introduced the concept of disentanglement via structured sparsity between the latent codes and the group generators. 
However, oi-VAE only factorizes the model on the generative side, leaving the inference network oblivious to the multi-aspect nature of the data. This makes handling missing data impossible, and leads to a less elegant approach when faced with different modalities that require their own inference network architectures.



Beyond neural networks, modeling multi-aspect and multi-modal data has been considered in a number of previous works. For instance, manifold relevance determination (MRD)~\citep{damianou2012manifold} attacks a similar problem, but in the context of Gaussian processes latent variable models~\citep{lawrence2004gaussian}. Specifically MRD learns group-specific (``private'') subspaces of the latent space and a global (``shared'') subspace. However, no prior is placed on the actual weights associating these subspaces to each of the generative models meaning that although this ``private''/``shared'' behavior may arise, it is not directly encouraged and will not exactly prune dimensions for each group. Though the framework could theoretically extend beyond two groups, the authors only considered two in their work.  There are a number of important decisions to be made in extending this framework to more groups. We leave that to future work and focus on comparisons with VAE-based frameworks.

\section{Experiments}

\subsection{Bars simulated data}
\begin{figure}[t!]
\vspace{-0.1in}
    \centering
\begin{tabular}{cc}
	\hspace{-0.2in}\includegraphics[height=1.35in]{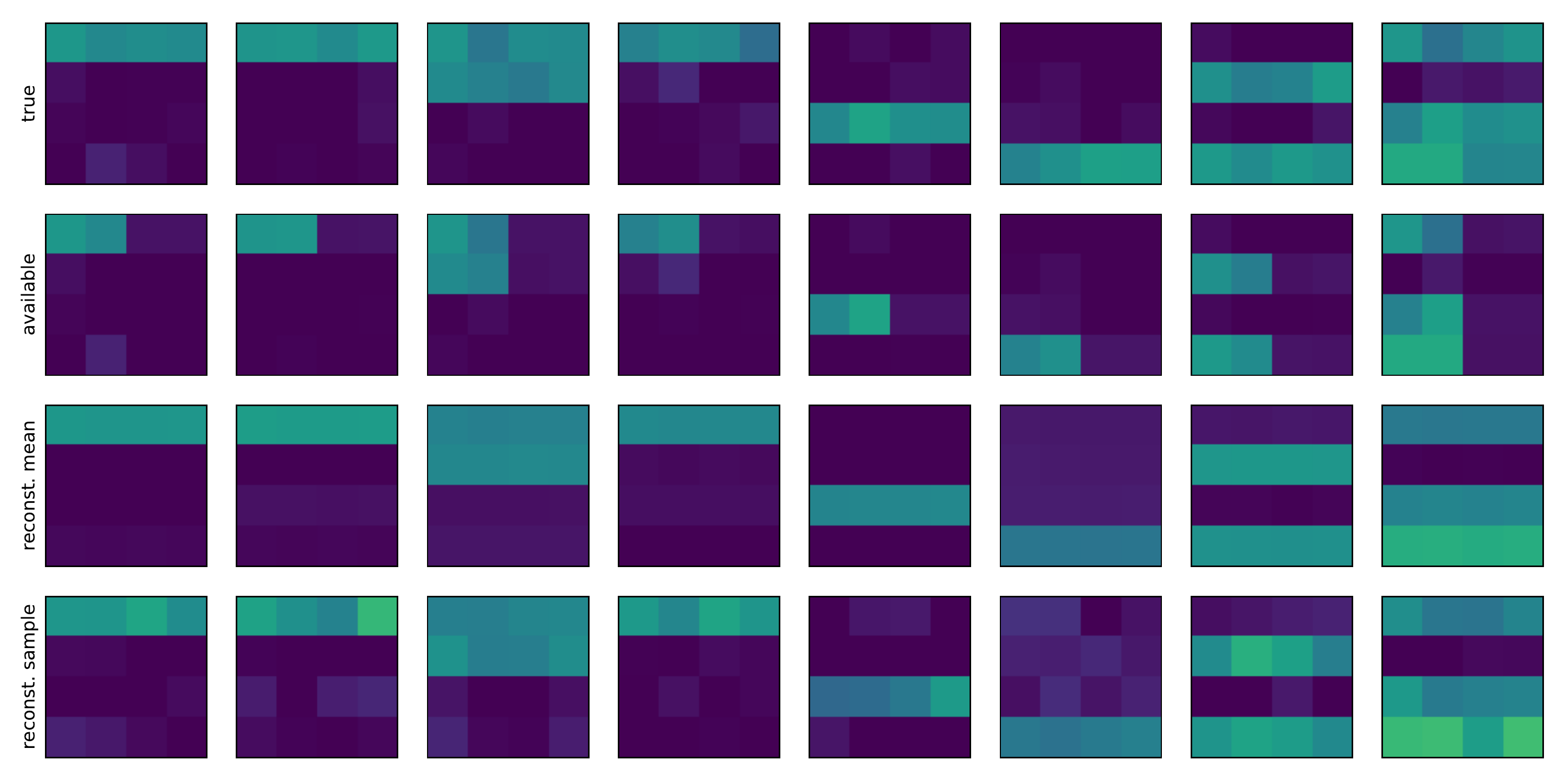} & \hspace{-0.1in}\includegraphics[height=1.4in]{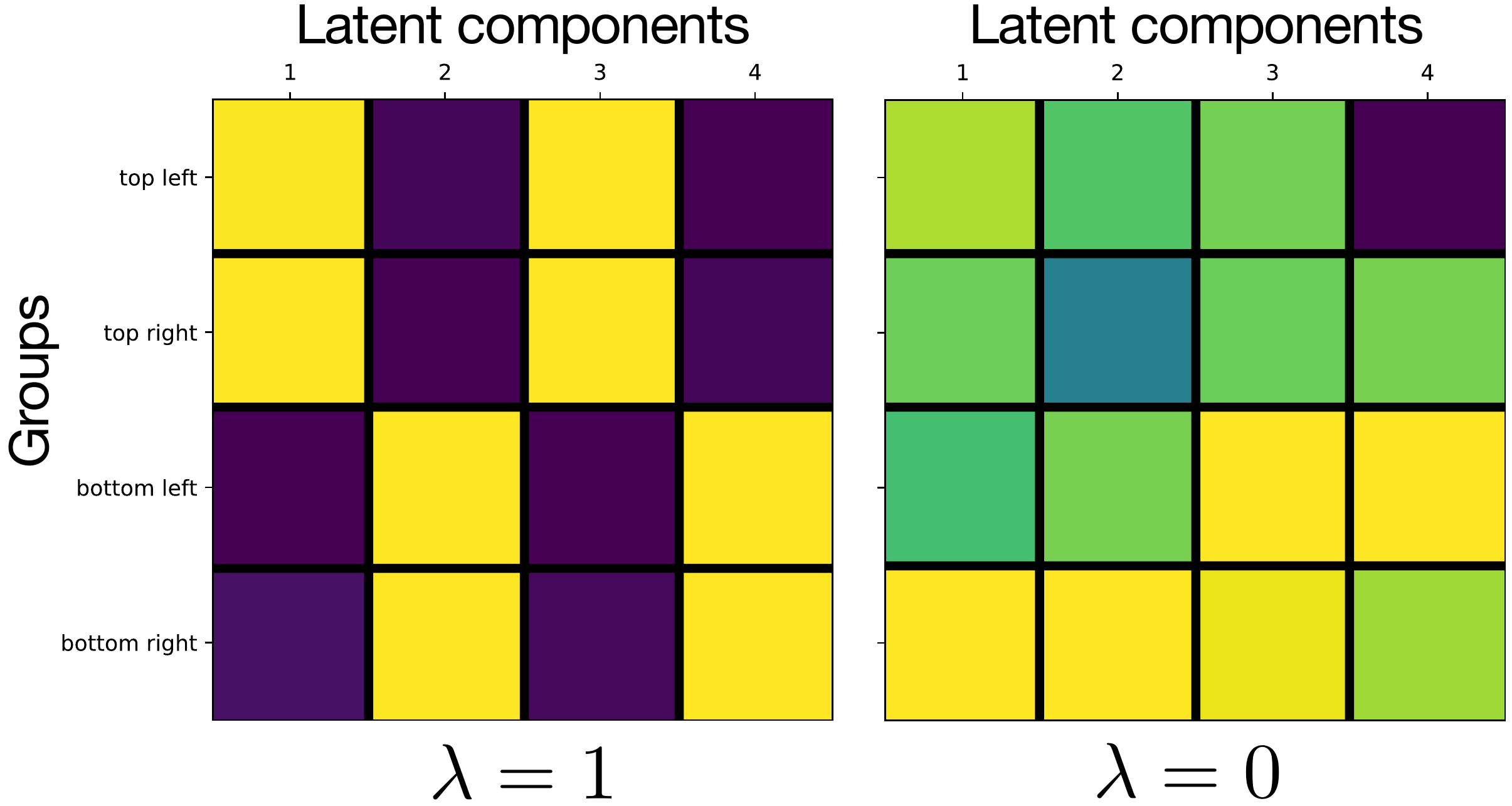}
\end{tabular}
\caption{\small \emph{Left:} A random subset of simulated images of horizontal bars plus noise (top row), training images with quadrants randomly removed (second row), reconstructions passing the variational mean (third row), reconstructions based on a randomly sampled latent code (bottom row). \emph{Right:} Learned factorization sparsity showing relationships between groups and latent dimensions.}
\vspace{-10pt}
\label{fig:bars}
\end{figure}

To assess the ability of our model to handle simple structured data with well understood correlations, we created synthetic images with rows of pixels randomly activated in each (see Figure~\ref{fig:bars}(\emph{left})). The image is split into its four quadrants, each one becoming a group. Since all the activity is horizontal, we should expect there to be latent components shared between the top two quadrants and the bottom two quadrants, but no shared components between the left two quadrants and the right two quadrants. In Figure \ref{fig:bars}(\emph{right}) we can see that this is exactly the case when $\lambda = 1$; in contrast, when $\lambda = 0$ (akin to a VAE with factored networks) the sparsity pattern disappears completely. We can also see that the model has learned to successfully reconstruct the right half of images when presented with only the left half based on its learned structure across groups in the model.

\subsection{Motion capture}
\begin{figure}[t!]
\vspace{-0.2in}
    \centering
	\hspace{-0.1in}\includegraphics[height=1.8in]{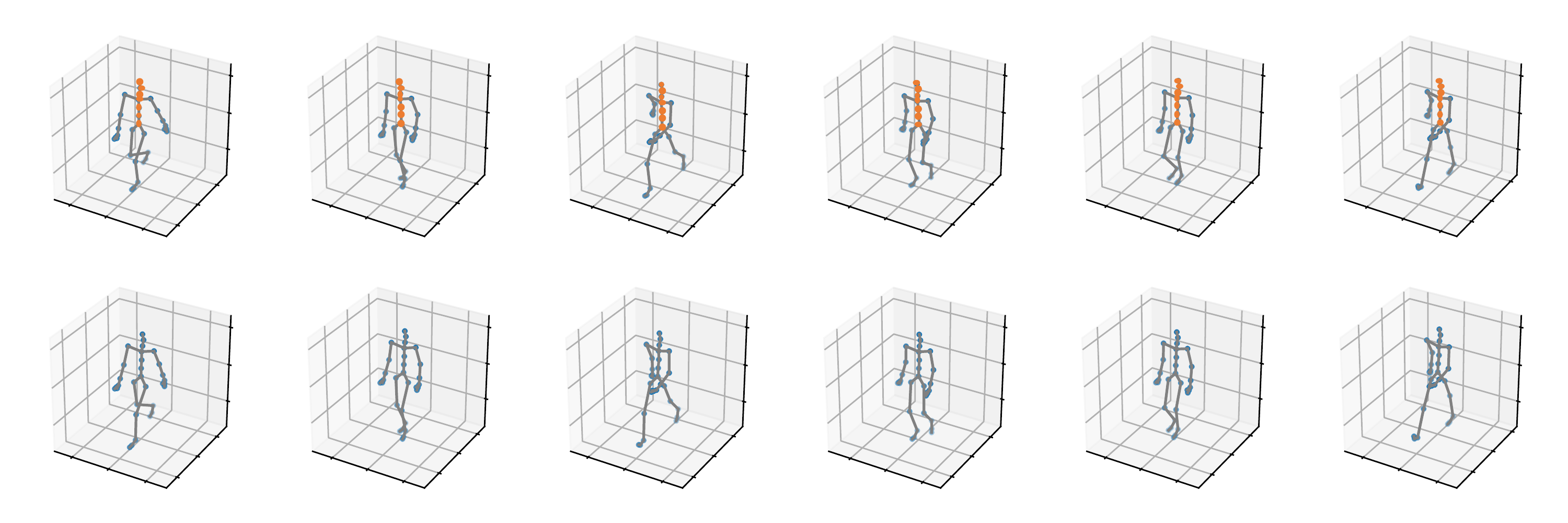}
\caption{\small Reconstructing poses from a subset of observed groups.  \emph{Top:} Input pose with provided joint measurements highlighted (core only). \emph{Bottom:} Reconstructed poses.}
\vspace{-16pt}
\label{fig:mocap_reconstructions}
\end{figure}

In order to assess factVAE's ability to recover interpretable sparsity in real-world data, we built and trained a factVAE model on motion capture data from CMU's motion capture database. The results of this model are presented in Figures \ref{fig:mocap_reconstructions} and \ref{fig:mocap_sparsity}. We divided the skeleton into 5 major groups: the back and head (``core''), right arm, left arm, right leg, and left leg.

In Figure \ref{fig:mocap_sparsity} we can clearly see that factVAE has successfully learned a sparse, disentangled representation for the data in which each latent component interacts with only a few of the groups. The exceptions to this are components 1 and 4 which we found encode the position in the overall stride of the subject's walk.  We quantitatively see the benefits of this learned representation in Table~\ref{table:mocap_heldoutloglike}, where we compute heldout log likelihood of a walking sequence, comparing to our model \emph{without} sparsity, as well as to the JMVAE+ model described in Section~\ref{section:related-work}.  The benefits of our framework are clear, but especially for limited training data scenarios.

In Figure \ref{fig:mocap_reconstructions} we show reconstructions conditioning only on joints in the core, including the head and back. Although only subtle motions can be seen visually, factVAE has successfully learned that rotations in the spine and head are correlated with the remaining groups and can be used to produce very accurate reconstructions of the overall pose.
%


\begin{minipage}{\textwidth}
  \begin{minipage}[b]{0.49\textwidth}
    \centering
    \includegraphics[height=1.5in]{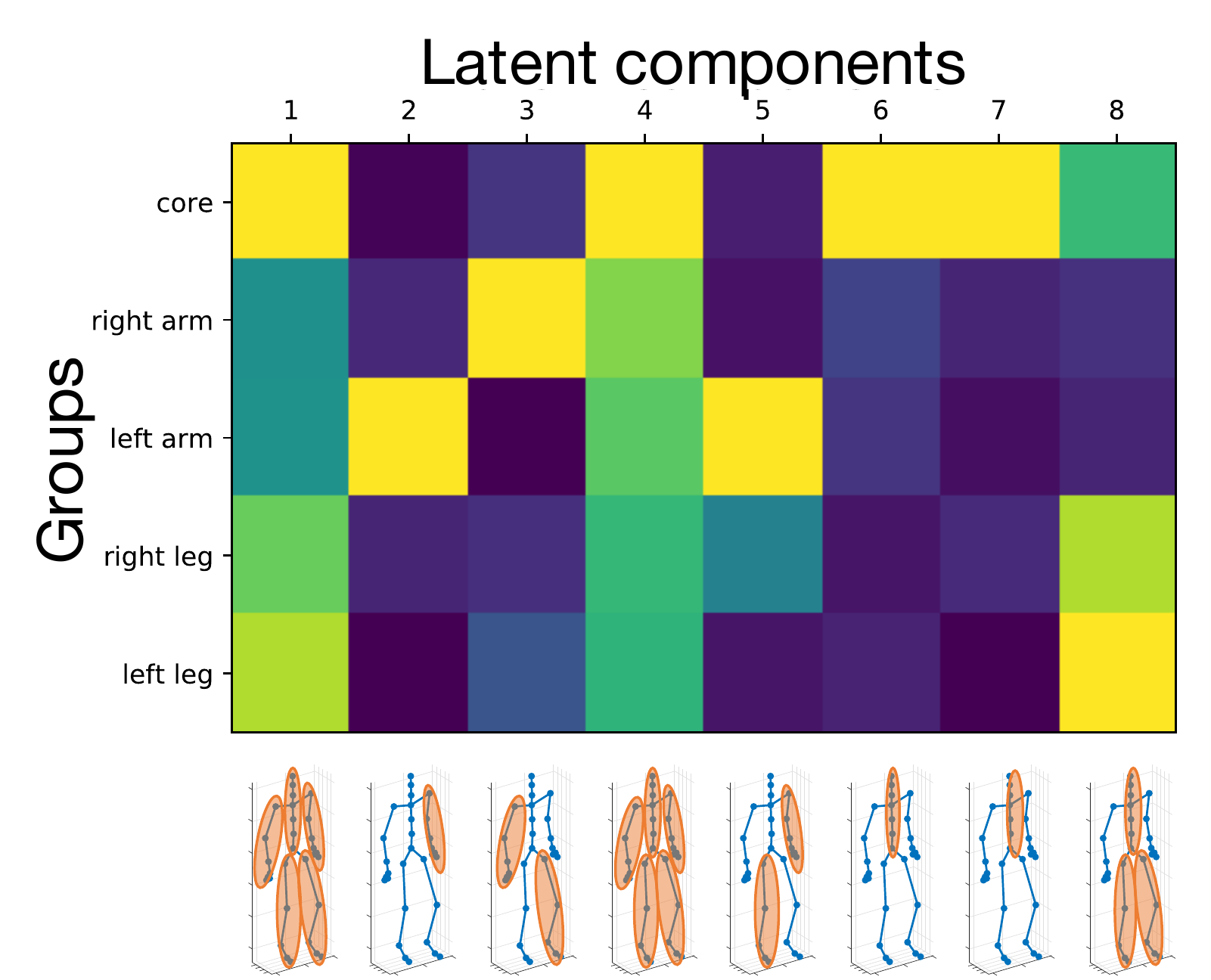}
    \captionof{figure}{\small Learned factorization sparsity showing relationships between groups (limbs and core) and latent dimensions.  Highlighted groups on skeletons correspond to most prominent groups connected to a given latent dimension.}\label{fig:mocap_sparsity}
  \end{minipage}
  \hfill
  \begin{minipage}[b]{0.49\textwidth}
    \centering
    \begin{tabular}{|l|cccc|}
		\hline
		Trials & 1 & 2 & 5 & 10 \\
		\hline
		JMVAE+ & $-620$ & $-166$ & $-77$ & $54$ \\
		$\lambda = 0$ & $-600$ & $-117$ & $-21$ & $\mathbf{89}$\\
		$\lambda = 1$ & $\mathbf{-555}$ & $\mathbf{9}$ & $\mathbf{-13}$ & $83$ \\
		\hline
	\end{tabular}
	\vspace{0.65in}
      \captionof{table}{\small Test log-likelihood of a heldout mocap walking sequence, varying the number of training sequences (trials), comparing factVAE with sparsity ($\lambda=1$) and without ($\lambda=0$).}\label{table:mocap_heldoutloglike}
    \end{minipage}
  \end{minipage}




\subsection{Faceback \protect\footnote{``You take a picture of anybody's face, and it'll show you what the back of his head looks like. Faceback!'' -- \textit{Will Ferrell and Eva Mendes in ``The Other Guys''}}}
\label{sec:faceback}

\begin{figure}[t!]
\vspace{-0.1in}
    \centering
	\includegraphics[width=0.75\columnwidth]{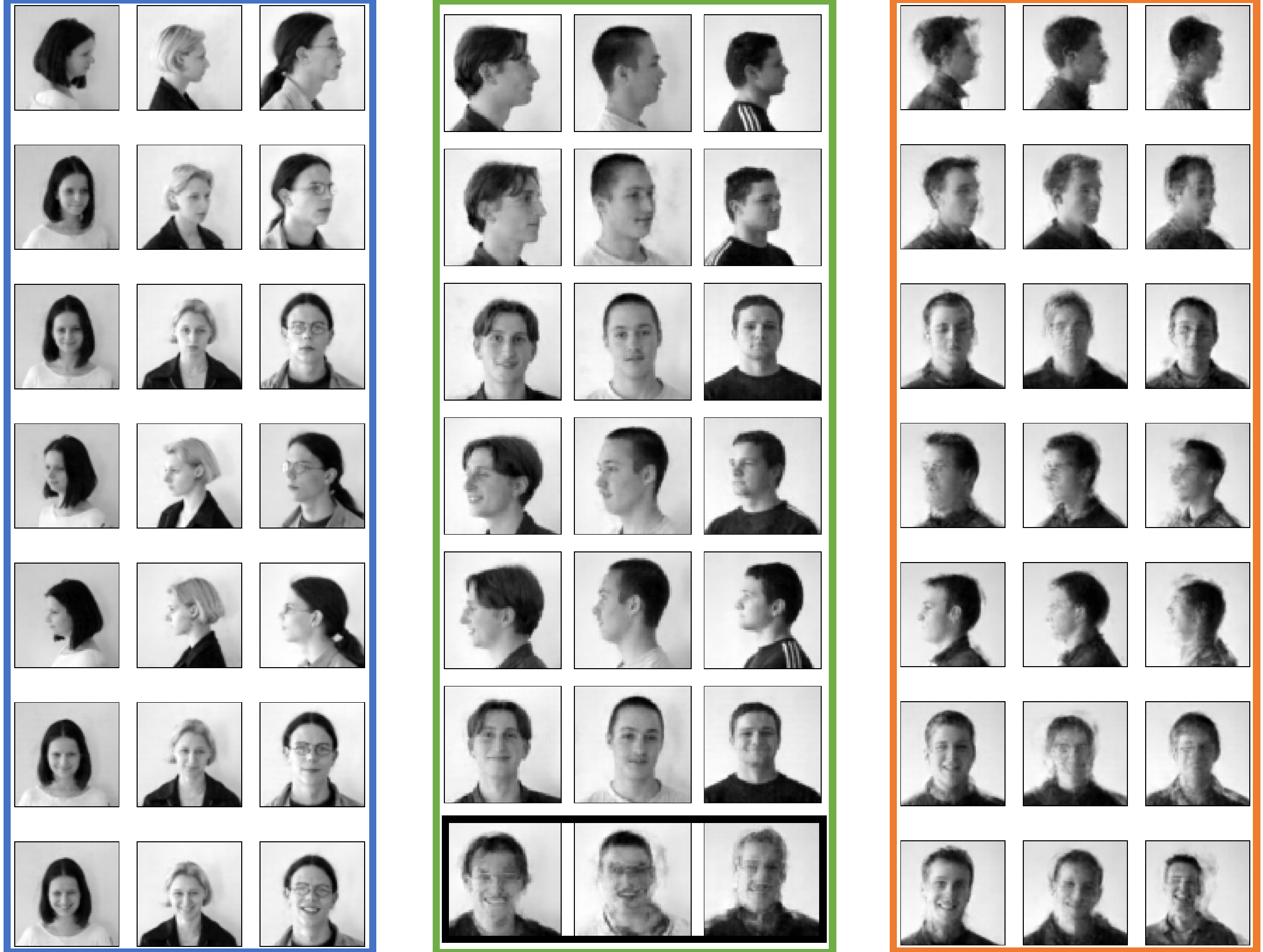}
\caption{\small Samples from the model. Each column corresponds to a subject. Each row shows a pose reconstructed from all \textit{other} available poses for that subject. All images are reconstructions. \textit{(blue, left)} Subjects present in the training set containing full data. \textit{(green, middle)} Subjects present in the training set but missing the smiling-with-teeth pose, highlighted in black. \textit{(orange, right)} Subjects in the heldout test set that are never seen at training time. The displayed subjects are the first ones in each set; no cherry picking was performed.}
\vspace{-10pt}
\label{fig:faceback}
\end{figure}

In order to further test factVAE's ability to handle rich, complex data across many different views, we trained and evaluated factVAE on a face reconstruction task. We gathered images from the CVL Face Database (\url{http://www.lrv.fri.uni-lj.si/facedb.html}) and treated each view as a group, and each subject as an example. The dataset contains images of 114 subjects presented in 7 different poses: 5 rotational, and 2 smiling. This dataset is especially interesting since it actually contains missing data; not all subjects have images for all 7 poses. We use the first 100 subjects for training and the remainder for testing. We used an adaptation of the DCGAN~\citep{radford2015unsupervised} architecture for the inference and generative networks. See the Supplement for more details. Results are shown in Figure \ref{fig:faceback}.

We evaluated the model by reconstructing each of the 7 views based on the \textit{other available views}. All of the results presented are reconstructed images. As we can see, the factVAE is able to clearly reconstruct the training images (blue box) with no trouble, and it successfully picks up on salient features even on completely unseen inputs (orange box). For example, the final row shows the smiling-with-teeth group, and we can see clearly that the model is able to generate smiling-with-teeth images even for completely unseen subjects.  We also include reconstructions for some subjects where the open-mouth smiling image was missing in the CVL Face Database (orange box).  Though they are noisy, the model clearly captures smiling with teeth.\footnote{The faces in this dataset were not centered in the images, leading to another source of reconstruction error.}

\section{Discussion}
We proposed factVAE, a deep generative model that can handle multi-aspect data and is robust to training and reconstructing with missing data. Traditional deep generative models either cannot handle missing data in the first place or only partially address the issue of reconstruction with arbitrary conditioning. On the other hand, factVAE addresses the problem from the ground up, factorizing both the inference and generative networks across groups and coherently aggregating information through a product of experts approximate posterior. We also incorporate shared sparsity in order to disentangle latent components and groups. As a consequence, we can apply factVAE successfully even in limited data scenarios and extract interpretable information.

We demonstrated that factVAE is able to recover true sparsity in our synthetic bars experiment. On real-world data, including motion capture sequences and pictures of faces, we found that factVAE performed quantitatively superior to other techniques in reconstruction metrics and qualitatively produces realistic samples even when conditioning on as little as a single group. Investigating the learned sparse group-latent component relationships, we found that factVAE produces interpretable and meaningful modes of variation.
Encouraged by these results, we are excited for further exploration of structure and sparsity in deep generative models, and investigating their impact in other application domains.

\pagebreak
\bibliographystyle{unsrt}
\bibliography{main}

\end{document}